%% file: main.tex
\documentclass[twoside,leqno,twocolumn]{article}

\usepackage[letterpaper]{geometry}
\usepackage{inputenc}
\usepackage{ltexpprt}
\usepackage[english]{babel}
\usepackage{hyphenat}
\usepackage{graphicx} 
\usepackage{csquotes}
\usepackage{amssymb}
\usepackage{amsfonts}
\usepackage{amsmath}
\usepackage{url}
\usepackage{blindtext}
\usepackage{microtype}
\usepackage{booktabs}
\usepackage{mathrsfs}
\usepackage{algorithm}
\usepackage{algorithmic}
\usepackage{booktabs}
\usepackage{multirow,adjustbox}
\allowhyphens
\makeatletter
\newcommand{\printfnsymbol}[1]{%
  \textsuperscript{\@fnsymbol{#1}}%
}
\makeatother

\newcommand{\cha}{\textsc{Chameleon}}
\newcommand{\leo}{\textsc{LEO}}
\newcommand{\hydra}{\textsc{hidra}}

\newcommand{\maml}{\textsc{maml}}
\newcommand{\reptile}{\textsc{reptile}}

\DeclareMathOperator* {\argmin}   {arg min}
\newcommand {\train}  {{\text{train}}}
\newcommand {\test}   {{\text{test}}}

\newcommand {\N}   {{\mathbb{N}}}
\newcommand {\T}   {{\mathcal T}}

\begin{document}

\title{\Large \textsc{HIDRA}:  Head Initialization across Dynamic targets for Robust Architectures} 
\author{Rafael Rego Drumond\thanks{University of Hildesheim, Information Systems and Machine Learning Lab (\{radrumond,brinkmeyer,schmidt-thieme\}@ismll.uni-hildehseim.de)} \thanks{Equal Contribution}
\and Lukas Brinkmeyer\printfnsymbol{1}\printfnsymbol{2} \and Josif Grabocka\printfnsymbol{1} \and Lars Schmidt-Thieme\printfnsymbol{1}}

\date{}

\maketitle


\fancyfoot[R]{\scriptsize{Copyright \textcopyright\ 2020 by SIAM\\
Unauthorized reproduction of this article is prohibited}}





\begin{abstract} \small\baselineskip=9pt 
\input{chapters/0-abstract.tex}
\end{abstract}

\input{chapters/1-intro.tex}

\input{chapters/2-related.tex}
\input{chapters/4-method.tex}
\input{chapters/5-exp.tex}

\input{chapters/6-conclusion.tex}
\bibliographystyle{acm}
\bibliography{bib}
\end{document}

%% file: chapters/0-abstract.tex
The performance of gradient-based optimization strategies depends heavily on the initial weights of the parametric model. Recent works show that there exist weight initializations from which optimization procedures can find the task-specific parameters faster than from uniformly random initializations and that such a weight initialization can be learned by optimizing a specific model architecture across similar tasks via \maml{} (Model-Agnostic Meta-Learning). Current methods are limited to populations of classification tasks that share the same number of classes due to the static model architectures used during meta-learning. 
In this paper, we present \hydra, a meta-learning approach that enables training and evaluating across tasks with any number of target variables. We show that Model-Agnostic Meta-Learning trains a distribution for all the neurons in the output layer and a specific weight initialization for the ones in the hidden layers. \hydra{} explores this by learning one master neuron, which is used to initialize any number of output neurons for a new task. Extensive experiments on the Miniimagenet and Omniglot data sets demonstrate that \hydra{} improves over standard approaches while generalizing to tasks with any number of target variables. Moreover, our approach is shown to robustify low-capacity models in learning across complex tasks with a high number of classes for which regular MAML fails to learn any feasible initialization.

%% file: chapters/1-intro.tex
\section{Introduction} \label{sec:intro}


\input{plots/plotplots/testfig}

Machine learning models and especially deep neural networks are crucial in various fields of research and industry to the point that not only experts but also practitioners of related areas are dependent on their application. In almost all cases, the optimization of these parametric models relies on a suitable selection of multiple hyperparameters, which influence the training performance drastically. This parameter selection either requires expert knowledge or the use of hyperparameter optimization techniques \cite{schilling2016scalable}. 
One often disregarded hyperparameter is the weight initialization for parametric models, which is required as a starting point for gradient-based-optimization. A suitable weight initialization is essential for a fast convergence to a near-optimal solution when using a method that generally converges to a local optimum. Standard hyperparameter optimization approaches are not capable of finding a per-weight initialization for neural networks due to their high number of continuous weight parameters. Instead, a random weight-initialization is typically chosen as a starting point \cite{glorot2010understanding,He_2015_ICCV}. 

Recent approaches such as \maml{} \cite{finn2017model} show that it is possible to learn a weight initialization for a specific neural network by utilizing second-order optimization for training across a set of similar tasks. This allows us to find a per-weight initialization that can lead to a fast convergence for similar tasks. However, such a process requires that each task has the same number of target variables since a specific model architecture is optimized, which also means having a fixed number of output neurons. 
In practice, it results in a huge computational effort since it is necessary to optimize a single model architecture for each potential number of output neurons expected during application.
Moreover, the initialization should perform equally well for two identical tasks with permuted class-order due to the fact that there is no inherent sequence to the target variables of a standard classification task. This suggests that the different output neurons cannot learn different output weights when trained across data sets with different class semantics.
We propose an extension to existing meta-learning approaches by learning a single master neuron, which can be used to initialize any number of output neurons. During meta-learning, it is used to initialize the required number of output neurons for a specific task, train on the task via \maml{}, and update the master neuron with regard to the different output neurons.
Thus, enabling approaches like \maml{} to train and evaluate across tasks with a dynamic number of target variables.

The core contributions of this work are as follows:
(1) We demonstrate that standard \maml{} learns indifferent output neurons, which limits the approach to a fixed number of target variables.
(2) We extend \maml{} to work across dynamic target sizes by deploying a general master neuron that learns to initialize any number of output neurons for similar tasks.
(3) Finally, we show that our method \hydra{} leads to higher model robustness such that even for tasks with a high number of target variables, finding a suitable weight initialization is feasible while regular \maml{} fails to do so (Figure \ref{testfig}).


%% file: plots/plotplots/testfig.tex
\begin{figure}[ht]
\centering
\includegraphics[width=.85\columnwidth]{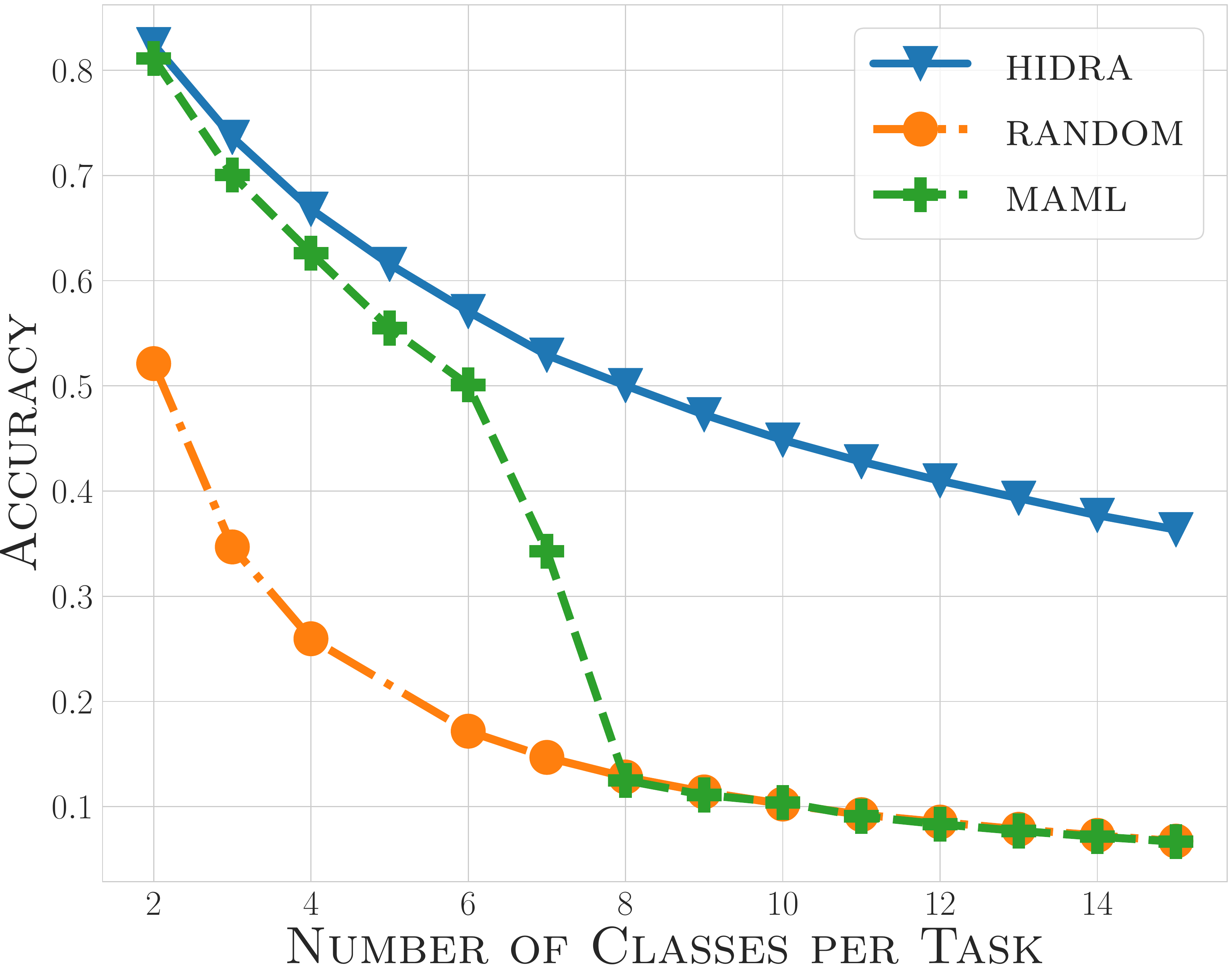}
\caption{Performance comparison of Random Initialization and \maml{} vs. a single initialization learned with \hydra{} for different number of target variables when training on Miniimagenet} \label{testfig}
\end{figure}

%% file: chapters/2-related.tex
\section{Related Work} \label{sec:related}

    Current meta-learning approaches that find a model-weight initialization are typically evaluated by applying them to \textit{few-shot classification problems} because it is generally easier to generate the necessary number of tasks required for meta-learning when dealing with few-shot tasks.
    Few-shot learning \cite{gui2018few, sung2018learning, Gidaris2018DynamicFV} strives to achieve the highest possible classification performance when faced with a new task that comprises only a handful of samples per class. This can be achieved by learning an initialization that converges fast, even when only a few instances are given, but also through the application of other meta-learning approaches. For example, Liu et al. \cite{Liu2018LearningTP} try to deal with this low-data regime by classifying all available tasks in one step by using transductive inference through label propagation as opposed to having a model that processes single tasks. Snell et al. \cite{snell2017prototypical} propose to learn a single model via meta-learning that embeds instances of a task in a metric space to measure the distances between them. For a novel task, a prototypical representation is selected for each target class to predict new images simply by looking at the nearest neighbor among these prototypes.
    To better calculate these distances, Oreshkin et al. devised TADAM \cite{oreshkin2018tadam}, a relation metric that adapts based on the task and scales appropriately as well.
    
    In contrast to these methods, there are many approaches that strive to optimize the model on the training instances of the evaluation task, instead of simply using them for inference. Training a model on a single few-shot task with such a small number of samples requires a suitable model initialization because it can very easily converge to a poor local optimum otherwise. 
    
    Another category of meta-learning approaches is referred to as \textit{Transfer Learning} \cite{sung2018comp,pan2010survey}. It describes the process of training a model on different auxiliary tasks and then using the learned model to actually fit the target problem to improve performance. For instance, pre-training blocks of convolutional neural networks on smaller tasks allows fitting a joint model to a much larger task \cite{zoph2018learning}. Another angle of transfer learning is using auxiliary tasks to help the model extract more useful features by training extra heads of the architecture to learn metadata from the same inputs \cite{ranjan2019hyperface}.
    
    Our work builds on the research of Finn et al. Model Agnostic Meta-Learning (\maml{}) \cite{finn2017model} finds an initialization for a specific model $M$ by training it across a set of similar few-shot learning tasks. The authors optimize a single model initialization $\theta^{*}$ by taking into account the validations loss after some iterations on each task after starting on this same initial point. Every task consists of pairs of inputs and target values. This means the authors sample a batch of tasks $\tau$ from a greater set $\T$.
    New parameters $\theta'_\tau$ are then calculated for a task ${\tau}$ after performing one or more update steps using the specific loss $\mathcal{L}_{\tau}$ starting with the initialization $\theta^{*}$. For the update steps, after initializing $\theta'_\tau \leftarrow \theta^{*}$, this can be written as:
    %
    \begin{equation}\label{ma1}
        \theta'_\tau \leftarrow \theta'_\tau - \alpha \nabla_{\theta'_\tau} \mathcal{L}_{\tau}^{train}(M_{\theta'_\tau})
    \end{equation}
    Then $\theta^{*}$ is updated using the second derivative of the updated weights with regard to their validation performance over all tasks in the meta-batch ${\tau}$ as in:
    \begin{equation} \label{ma2}
        \theta^{*} \leftarrow \theta^{*} -
        \beta \frac{1}{|\T|}
        \nabla_{\theta^{*}}
        \sum_{i}
        {\mathcal{L}_{\tau}}^{val} (M_{\theta'_{\tau}})        
    \end{equation}    
    \maml{} can be applied to any architecture, but out-of-the-box will only work on a fixed topology. Alex Nichol et al. developed \reptile{} \cite{reptile2018} in order to simplify the heavy computation of second derivatives from \maml{} by approximating Equation (\ref{ma2}) as:
    \begin{equation} \label{rep}
        \theta^{*} \leftarrow \theta^{*} - \beta   \frac{1}{|\T|} \sum_{\tau} (\theta'_\tau - \theta^{*})        
    \end{equation}
    
    Finn et al. later expands this work with Probabilistic Model-Agnostic Meta-Learning  \cite{Finn2018ProbabilisticMM} to learn a distribution for the model parameters by injecting Gaussian noise into the gradient descent steps.
    \input{figs/plotfigs/hidra1}
    Inspired by \maml{}, a recent paper by Rusu et al. called \leo{} \cite{rusu2018metalearning} proposes a method to sample network parameters of a model for few-shot learning. An additional encoder network takes a task as input and generates a latent embedding that consists of a mean and a standard deviation for each neuron to initialize. These distributions are used to sample the parameters for the respective neurons. During the training process, the latent representation is updated instead of the weights itself. They show the effectiveness of that approach by achieving state-of-the-art results for few-shot-classification. The complexity of the generated latent embedding depends on the number of neurons to initialize since the approach generates a weight distribution to sample from for each neuron. Due to this computational bottleneck, the authors only generate the weights for an output layer that is placed on top of a pre-trained deep residual network, which is used to generate task-embeddings that facilitate learning.

    So far, none of these methods are specifically designed to work across tasks with different input and target shapes. The work by Brinkmeyer and Drumond et al., \cha{} \cite{brinkmeyer2019chameleon}, targets the problem of meta-learning tasks having a variable input schema. The authors train a network that transforms different input schema of training batches to a fixed representation, enabling meta-learning methods such as \reptile{} to work with tasks with different input sizes by attaching this model to the target network's input layer. Similarly, Dataset2Vec from Jomaa et al. \cite{Jomaa2019Dataset2VecLD} extracts useful meta-features from different data sets to perform hyperparameter optimization. 
  
    Our work focuses on meta-learning over tasks with different target variables, being the first to our knowledge to directly explore such a problem.

%% file: figs/plotfigs/hidra1.tex
\begin{figure*}[t!]
\centering
\includegraphics[width=\textwidth]{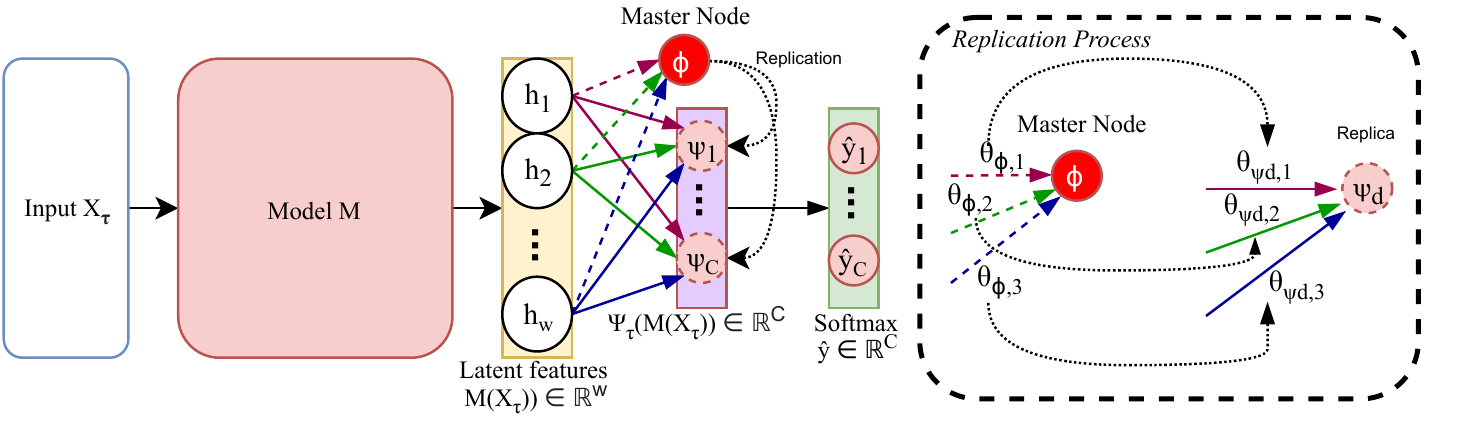}
\caption{\hydra{} utilizes a fixed Model $M$ but instead of a fixed output layer, the method keeps a single neuron $\phi$ parameterized with $\theta_\phi$. Given a task $\tau$ with $C$ target variables, $\phi$ is replicated $C$ times forming a task specific output layer $\Psi$ with neurons $\psi_d$ parameterized with $\theta_{\psi d}$. Dashed lines represent weights assigned to the master node with respect to the latent features from M. Dotted lines represent the master node replicating itself and its weights to create the output layer neurons.} \label{hidra}
\end{figure*}

%% file: chapters/4-method.tex
\section{Methodology} \label{sec:method}
Meta-learning approaches like \maml{} train a parameter initialization $\theta^{*}$ for a specific model $M$ by sampling meta-batches of similar tasks $\tau$ from a set of tasks $\T$. In few-shot classification, a task $\tau$ is represented by a single batch containing $K$ instances for each of the $N$ classes present. Thus, it consists of predictor data $X_{\tau}$ and target data $Y_{\tau}$.
Typically, similar tasks are defined to have the same feature space but different target variables such that $X_{i} \in \mathbb{R}^{NK \times F}$ and $Y_{\tau} \in \mathbb{R}^{NK \times N}$ with $F$ features for every task $\tau \in \T$. As usually defined in the literature, this type of problem setting is referred to as $N$-way-$K$-shot.
Thus, the goal is to provide a task with $K$ instances for each of the selected $N$ classes to the model and observe a high accuracy on unseen instances of that task after training. Furthermore, a task $\tau$ comes with a predefined training/test split $(X_{\tau}^{\train}, Y_{\tau}^{\train}, X_{\tau}^{\test}, Y_{\tau}^{\test})$

Most optimization-based meta-learning approaches operate in two phases: An inner loop and an outer loop. During the inner loop, the model $M$ is trained on a specific task starting from the current weight initialization $\theta^{*}$ for $U$ gradient steps. The updated parameter set $\theta'$ for a task $\tau$ is then given by:
\begin{equation}
    \theta_\tau' = G(M, \theta^{*}, steps, X_{\tau}^{\train}, Y_{\tau}^{\train})
\end{equation}
where $G$ is the optimization method used to compute the new weights by performing a number of inner update steps. For \maml{} this is shown in Equation \ref{ma1}.
Afterward, the performance of the current initialization is evaluated by measuring the validation performance for the same task with those updated weights. The outer loop executes the inner loop for a batch of tasks to update the current initialization $\theta^{*}$ with respect to the validation performance. The outer meta-objective is then defined as:
\begin{equation}\label{hi2}
     \argmin_{\theta^{*}} \mathbb{E}_{\tau \sim \T} \thinspace L_{val}(X_{\tau}^{\test}, Y_{\tau}^{\test} ,M, \theta_\tau') \\
\end{equation}
In \maml{} this outer objective is optimized by relying on the second derivatives as described in Equation \ref{ma2}.

Optimizing a fixed network architecture restricts the model $M$ to tasks with the same number of classes. As already stated previously, the learned initialization is required to be invariant to permutations of the class order since two sampled tasks could have the same instances $X$ while having their classes $Y$ in a different order. This means that there should be no inherent difference between the initialization learned for the different output neurons. At the same time, few-shot classification is always evaluated with unseen classes. Thus, it should be possible to learn a single output neuron initialization in the outer loop that can be dynamically adapted to each number of classes in the inner loop.

\subsection{HIDRA}

Our method learns a single output neuron as the master node $\phi$, which is replicated $c$ times during the inner-loop for a task with $c$ classes. \hydra{} takes into consideration that in every task, the number of classes might vary. Even when two tasks have the same amount of target variables, their labels may represent different classes. In essence, we need to find a dynamic architecture that works for any number of target variables while the initialization performs equally well for any label. In order to do so, we create the master node $\phi$. When $\phi$ is replicated $c$ times, it creates the output layer for a task that predicts $c$ number of values. This setup is illustrated in Figure \ref{hidra}.

Given a network architecture $M(x)$ with initial parameters $\theta^{*}_{M}$, we first randomly initialize the master node $\phi$ with parameters $\mathbb{R}^{w}$. In order to optimize the initialization, a batch of tasks with $C$ classes each is sampled. The number of classes only has to be identical within one meta batch and can vary over the course of the meta-training. During the inner loop, we generate a temporary output layer $\Psi$ with $C$ neurons each of which is initialized with the current weights of the master neuron $\phi$, so that the weights of the output layer $\Psi$ are set as
\input{algs/matrix}

This output layer is connected to the top of the base model $M$ to form the task-specific model $M_{\Psi}(x) = \Psi(M(x))$ which is capable of training on tasks with $C$ classes. 
Finally, we perform one meta-iteration of \maml{} on this model for some steps on each task $\tau$ of the sampled meta-batch to compute the task-dependent sets of weights ${\theta'_{M,\tau}}$ and ${\Psi'_{\tau}}$ using Equation \ref{ma1}, before updating the initial weights for $M$ and each neuron $\psi$ with Equation \ref{ma2}. Finally, we update the weights of the master neuron $\phi$ by aggregating the updated initial weights of the output neurons in $\Psi$:
\begin{equation}
    {\phi}^{new} = \frac{1}{C}\sum_{i}^{C} \psi_{i}
\end{equation}
The full approach is depicted in Algorithm 1. showing the inner and outer loop of \hydra{}.
\input{algs/hidra}

%% file: algs/matrix.tex
\begin{equation}\label{matmul}
    \psi_i = \begin{bmatrix} \theta_{i,1} & \theta_{i,2} & \dots & \theta_{i,w} \end{bmatrix} \\ 
\end{equation}
\begin{equation*}
    \Psi_{\tau} \leftarrow \phi =
    \underbrace{
    \begin{bmatrix}
        \theta_{1,1} & \dots  & \theta_{1,w} \\
         \vdots & \ddots & \vdots  \\
       \theta_{C,1} & \dots  & \theta_{C,w}
    \end{bmatrix}
    }_{\Psi_{\tau}^{\top}}   \leftarrow    \underbrace{
    \begin{bmatrix}
        \phi_{1} & \dots & \phi_{w} \\
         \vdots & \ddots & \vdots  \\
        \phi_{1} & \dots & \phi_{w}
    \end{bmatrix}
    }_{\phi \text{ replicas}}  
\end{equation*}

%% file: algs/hidra.tex
\begin{algorithm}[h]\label{alg:hydra}
\caption{\hydra{} Method}
\begin{algorithmic}[1]
    \STATE Select gradient step-sizes $\alpha$ and $\beta$
    \STATE Initialize Meta-Data-Set $\T$
    \STATE Initialize Model $M(x)$ with $\theta_{M}^{*}$
    \STATE Initialize Output Master Node $\phi$
    \FOR{Max-Iterations}
        \STATE Sample batch $b$ of tasks $\tau$ from $\T$ with a random\\
        ~ ~ ~ ~ ~ ~ ~ ~ ~ ~ ~ ~ ~ ~ ~ ~ ~ ~ ~ output size $c_{b} \thicksim \N$
        \STATE Instantiate output layer ${\Psi}_b$ with $c_{b}$ neurons
        \FOR{every neuron $\psi_k \in \Psi_b$}
            \STATE $\psi_{k} \leftarrow \phi$
        \ENDFOR
        \STATE Define network $M_{{\Psi_b}}(x) := \Psi_b(M(x))$
        \STATE $\theta_b = [\theta^{*}_{M} , \Psi_{b}]$
        \FOR{every task $\tau \in b$}
            \STATE $\theta_{\tau}'\leftarrow \theta_{b}$
                \FOR{$n$ amount of inner steps}
                    \STATE $\theta_{\tau}' \leftarrow$
                    $\theta_{\tau}'  - \alpha \nabla_{\theta_{b}} \mathcal{L}_{\tau}^{train}(M_{{\Psi_b}},\theta_{\tau}')$
                \ENDFOR 
        \ENDFOR
        \STATE $\theta_{b} \leftarrow \theta_{b} $
        $ - \beta \frac{1}{|b|}
        \nabla_{\theta_{b}}
        \sum_{\tau}
        {\mathcal{L}_{\tau}}^{val} (M_{\Psi_{b}}, \theta_{\tau}') $
        \STATE ${\phi} = \frac{1}{c_b}\sum_{\psi}^{\Psi_b} {\psi}$
    \ENDFOR
\end{algorithmic}
\end{algorithm}

%% file: chapters/5-exp.tex
\input{plots/plotplots/heatmap.tex}
\section{Experiments} \label{sec:exp}

We conduct experiments on the standard few-shot classification data sets Omniglot \cite{lake2011one} and Miniimagenet \cite{ravi2016optimization}. Both are frequently used as few-shot classification benchmarks. Omniglot consists of 1623 written characters, each with 20 instances, taken from 50 different alphabets. We randomly split the data set with 1200 characters used for training and the rest for testing. The Miniimagenet data set includes 100 classes from ImageNet with 600 instances per class. We utilize the proposed split with 64 classes in training, 16 in the validation, and 20 in the test data set as proposed by Ravi et al. \cite{ravi2016optimization}.

\input{plots/plotplots/graphsMINI.tex}
\input{plots/plotplots/OmniComp2}
\input{plots/plotplots/OmniComp1}
For all experiments, the same model architecture, originally proposed by \cite{vinyals2016matching}, and the same hyperparameters are used as in \cite{finn2017model}. It consists of four convolutional blocks, each being a 3x3 convolution, followed by batch normalization, ReLU nonlinearity, and 2x2 max pooling. For Omniglot, the filter size is set to 64 and for Miniimagenet to 32. The inner learning rate $\alpha$ for training the model on a specific task is set to 0.01 and 0.4 for Miniimagenet and Omniglot, respectively. For the meta-objective in Equation \ref{hi2}, the Adam optimizer \cite{adam} is used with a learning rate $\beta=1e^{-3}$. Our work focuses on $N$-way 5-shot classification tasks since this work focuses on analyzing varying class numbers. As for training, the number of inner gradient steps on a task is set to 5 and 1 for training Miniimagenet and Omniglot, respectively. Furthermore, for every meta-epoch, we sample 32 tasks for Omniglot and 4 for Miniimagenet. In contrast to the work of Finn et al. \cite{finn2017model}, we conducted the Omniglot experiments without data augmentation for \maml{} and \hydra{} which leads to a slightly lower accuracy but faster runtime to evaluate all the different number of classes. 
For evaluation, we aggregate the accuracy across 4000 randomly sampled test tasks, performing up to 10 gradients steps for Miniimagenet and accordingly up to 3 gradient steps for Omniglot on the learned initialization. We had to use an alternative implementation of \maml{} due to hardware scalability problems of the original implementation when evaluating tasks with a high number of classes. Running the original code for 2 to 6 classes per task leads to an approximately $5-6\%$ higher accuracy compared to the results reported in this work. Since we built \hydra{} on top of the same framework, we can assume that these findings transfer to other meta-learning approaches used for model initialization, including \maml{}.

\subsection{MAML}
In our first experiment, we want to analyze the weight initializations for the output layer learned via \maml{} to show there is no inherent structure between the neurons to motivate the application of \hydra. For that, we compared the performance of an initialization learned with \maml{} for $10$-way 5-shot Omniglot with the same initialization, but one of the ten learned output neurons is used to initialize every other output neuron. The results, shown in Figure \ref{weights1}, illustrate that the weights learned for a single output neuron with \maml{} are already sufficient to initialize the complete output layer. The average accuracy across each of those initializations is $94.49\%$, while the standard initialization using all learned output weights achieves $94.58\%$. Most importantly, using a single output neuron to initialize the output layer leads to higher performance in some of the cases. By visualizing these weights in Figure. \ref{weights2}, one can see how the output neurons are all learning a similar pattern showing the redundancy for the weights in the output layer learned via \maml{} (contribution 1).

\subsection{HIDRA}

For our main experiments, we compare the performance of \hydra{} and \maml{} when training on Omniglot and Miniimagenet for a varying number of classes. Our experiments investigate different $N$-way $5$-shot problem settings where the $N$ ranges from $2$ to $10$ classes for Omniglot and 2 to 15 classes for Miniimagenet. In order to compare our approach to \maml{}, which only works with a fixed number of classes, we train and evaluate a separate model with \maml{} for each output size $N$ as a baseline. Additionally, the performance of a standard random initialization is tracked for each class size. Finally, we learn various initialization with \hydra{} for two different settings: (i) with a fixed number of target variables and (ii) with a varying number of target variables utilized during meta-learning. Every initialization learned via \hydra{} is then evaluated for each of the different $N$-way settings. 
\input{plots/plotplots/plotGrid}
\input{tables/resultsMini}
\input{tables/resultsOmni}
\subsection{Results and Discussion}

The results of the comparison between the different initialization learned with \hydra{} on Miniimagent and the ones learned via \maml{} are shown in Figure \ref{minigraphs}. Note that each data point in the graph for \maml{} represents a separate model trained for the respective number of classes, while each graph of the \hydra{} experiments represents one model which is evaluated for every number of classes. The results for \hydra{} show a similar performance as \maml{} for up to 6 classes per task, with the \hydra{} models trained on 2-way and 10-way problems slightly outperforming \maml{}. Generally, all models initialized with \hydra{} generalize to any number of classes during evaluation (contribution 2). The results for training on tasks with varying target sizes achieve the highest accuracy with a slight improvement over the \hydra{} initialization trained for 2-way 5-shot classification.

Furthermore, \maml{} fails to generalize to unseen tasks when evaluating for more than 7 classes with 5 instances each when using the architecture described above. The performance of models initialized with \hydra{}, on the other hand, only decreases marginally with an increasing number of classes. Moreover, meta-learning with \hydra{} on tasks with a low number of classes is demonstrated to generalize to those with a high number of classes and vice versa, essentially computing a robust initialization that is independent of the number of target variables (contribution 3). The numerical results for these experiments are given in Table \ref{miexp}.

Experiments for evaluating our approach with varying number of gradient steps on Omniglot are displayed in Figure \ref{omnicomp2}. \hydra{} fails to outperform \maml{} with three gradient steps of size 0.4, as used in \cite{finn2017model}. In contrast to \maml{}, which reaches its highest accuracy after 3 steps, \hydra{} enables an even faster convergence and already overfits with three gradient steps on an unseen task, achieving the highest score after using a single step. Due to this faster convergence, we also evaluated \hydra{} with a smaller inner learning rate of 0.01 (Figure. \ref{omnicomp1}), which shows the best performance on Omniglot when using 3 gradient steps. Numerical results for Omniglot are displayed in Table \ref{omexp}.
The meta-learning progress for training \hydra{} on Omniglot for different $N$-way 5-shot settings is illustrated in Figure \ref{gridplot}. One can see that the gap between the training and validation error grows with the number of classes per task. The experiments were conducted using an NVIDIA Tesla K80. Training on Miniimagenet takes approximately 3.5 hours for 2 classes and 13 hours for 10 classes for both \hydra{} and \maml{}.  Our code is available online for reproduction purposes at https://github.com/radrumond/hidra.

%% file: plots/plotplots/heatmap.tex
\begin{figure}[!ht]
\centering
\includegraphics[width=\columnwidth]{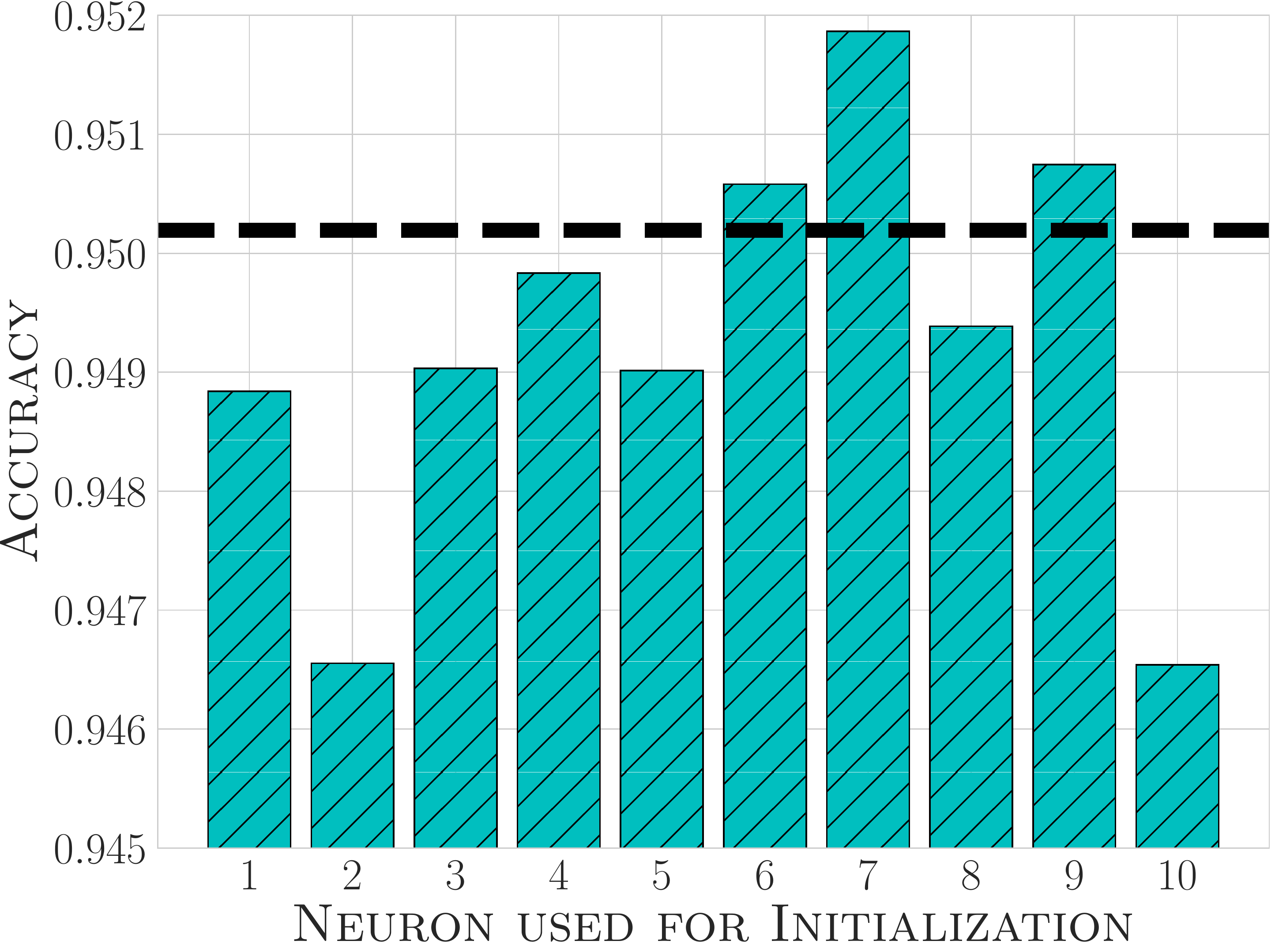}
\caption{Test accuracy for Omniglot 10-way 5-shot when using the weights for one output neuron learned  via \maml{} to initialize the other output neurons. The dashed line marks the performance of the regular initialization of \maml{}.} 
\label{weights1}
\end{figure}

\begin{figure}[ht!]
\centering
\includegraphics[width=\columnwidth]{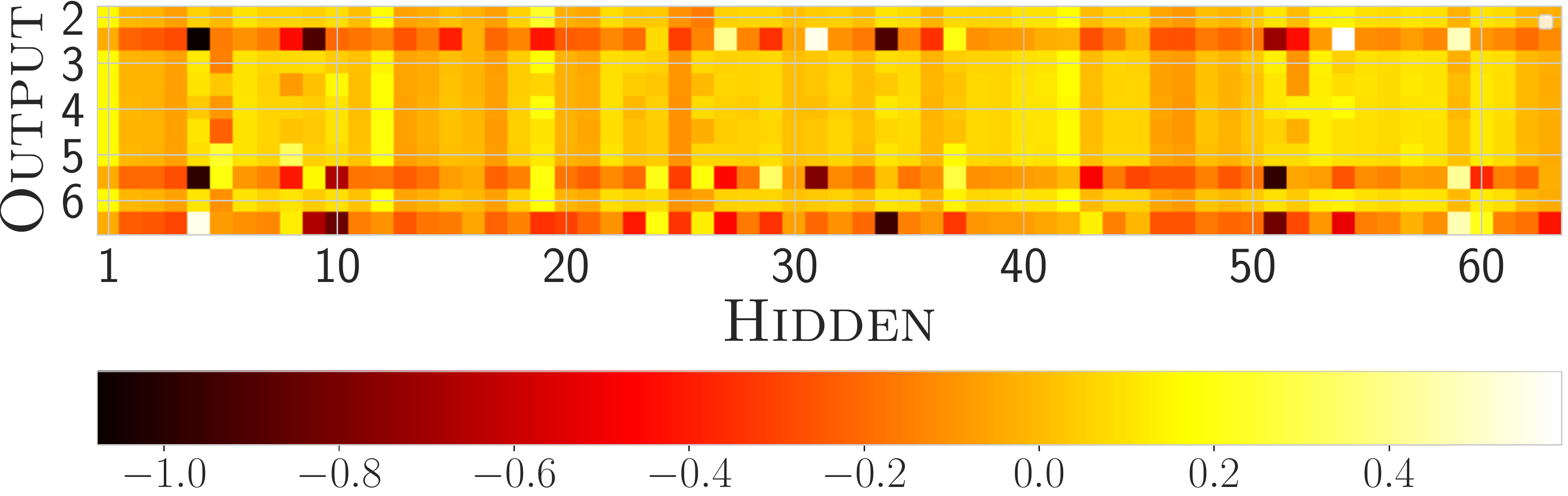} 
\caption{Weights of the output layer for the initialization learned via \maml{} when trained on Omniglot $10$-way.} \label{weights2}
\end{figure}

%% file: plots/plotplots/graphsMINI.tex
\begin{figure*}[ht!]
\centering
\includegraphics[width=.49 \textwidth]{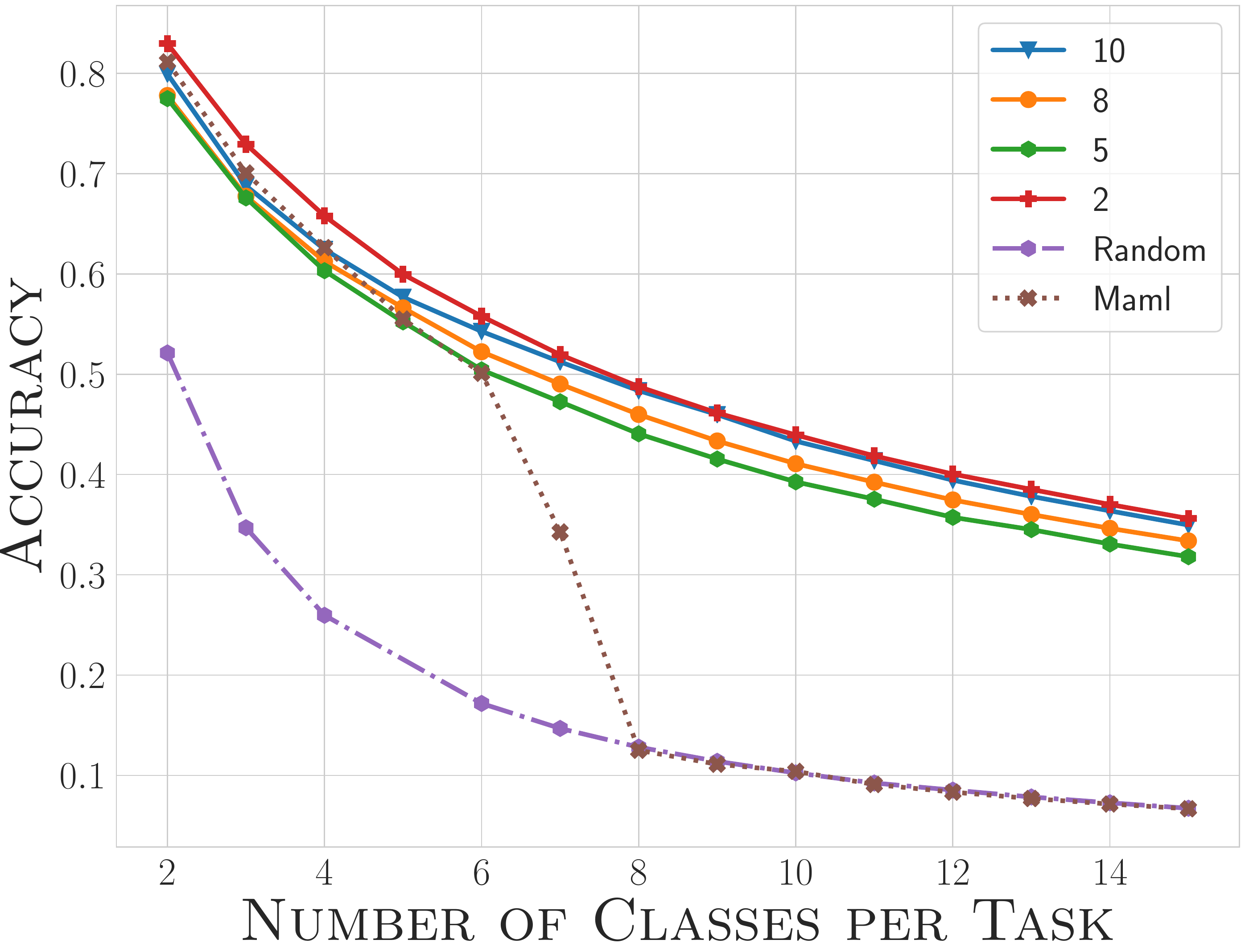} 
\includegraphics[width=.49 \textwidth]{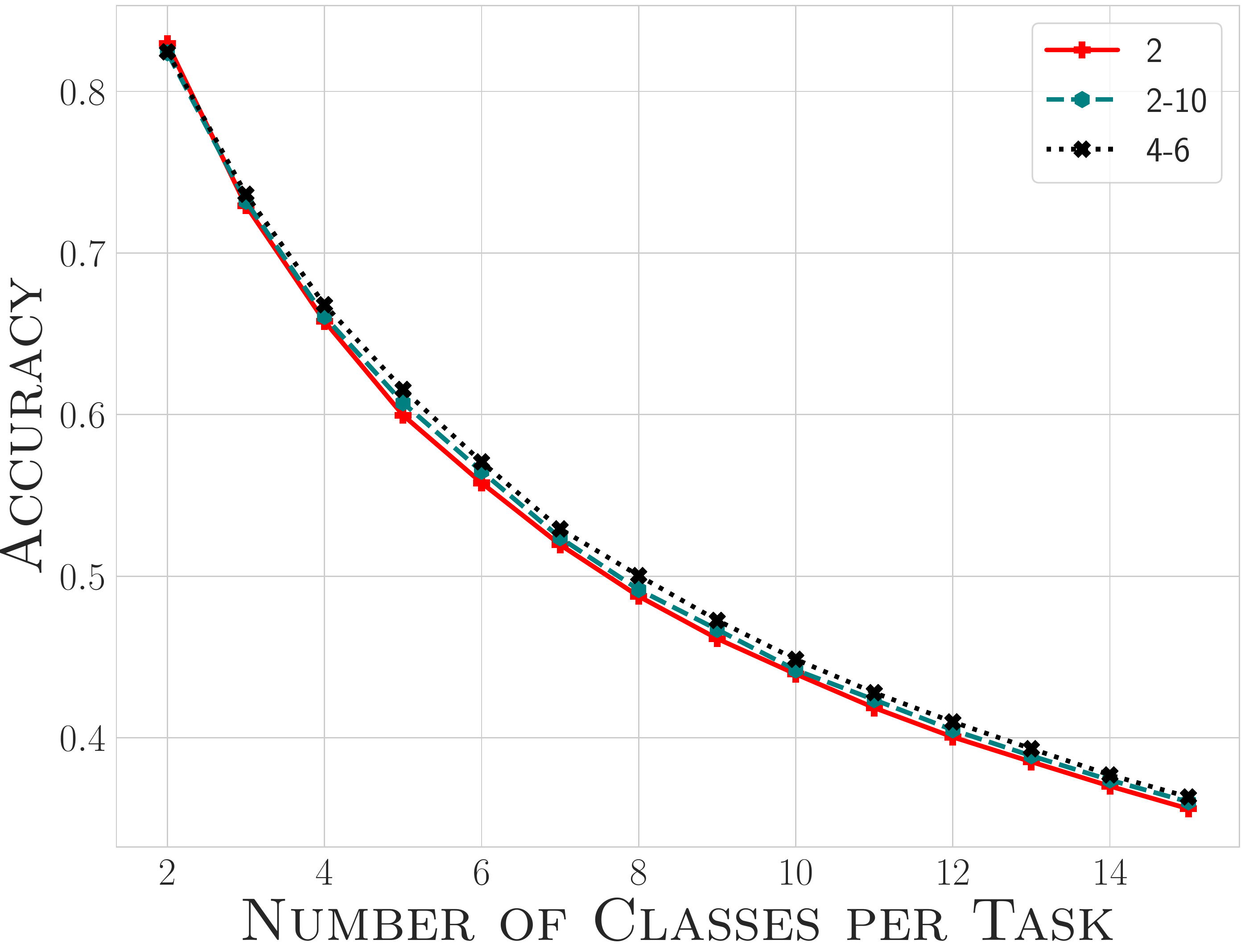}
\caption{Performance comparison between \maml{} and \hydra{} for different numbers of target variables when training on Miniimagenet. The left plot compares \hydra{} trained with static number of target variables (i). The right plot compares the best result from the static experiments with dynamic experiments where we have varied sizes of target variables within a range (ii).} 
\label{minigraphs}
\end{figure*}

%% file: plots/plotplots/OmniComp2.tex
\begin{figure*}[ht!]
\centering
\includegraphics[width=.32\textwidth]{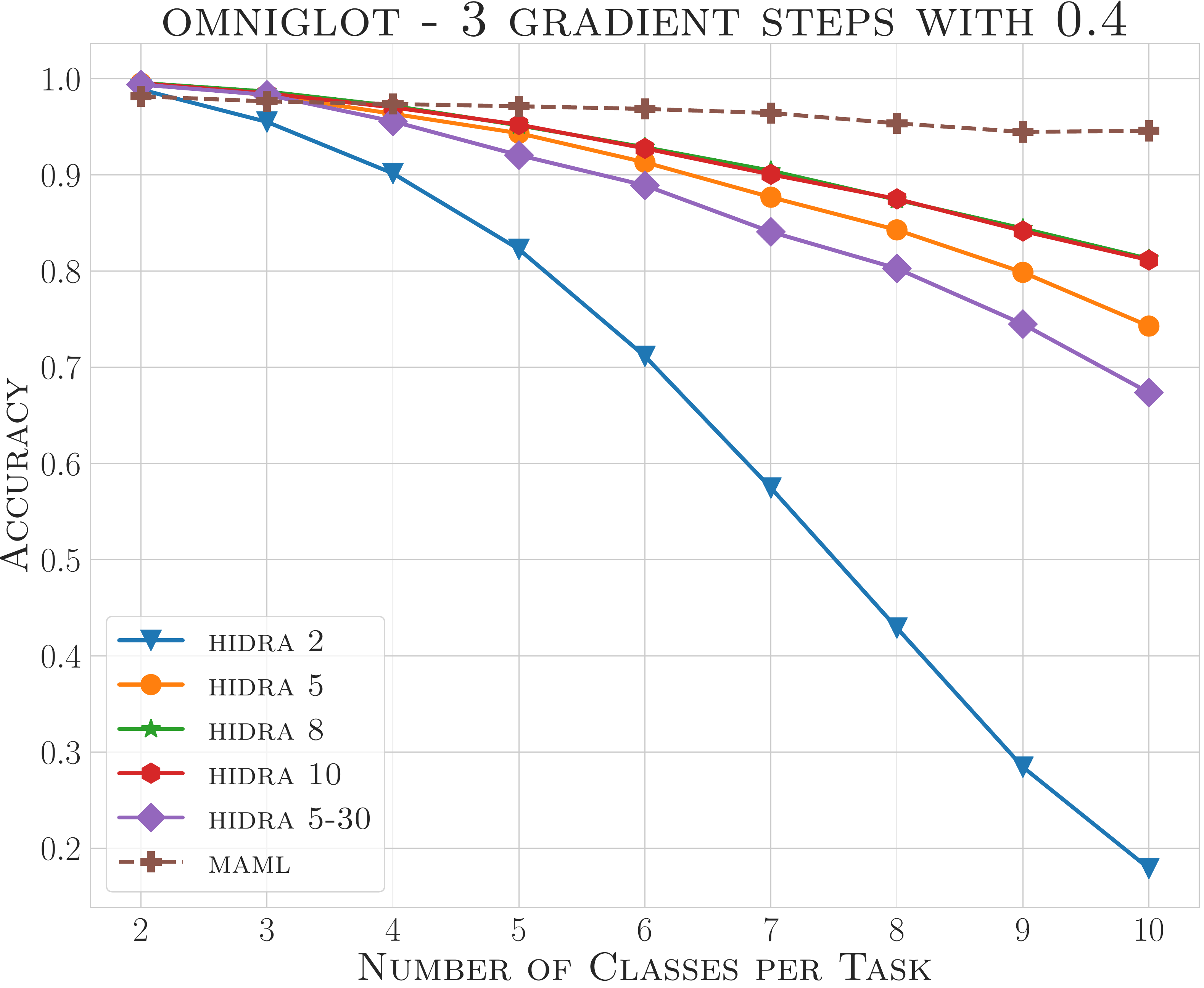}
\includegraphics[width=.32\textwidth]{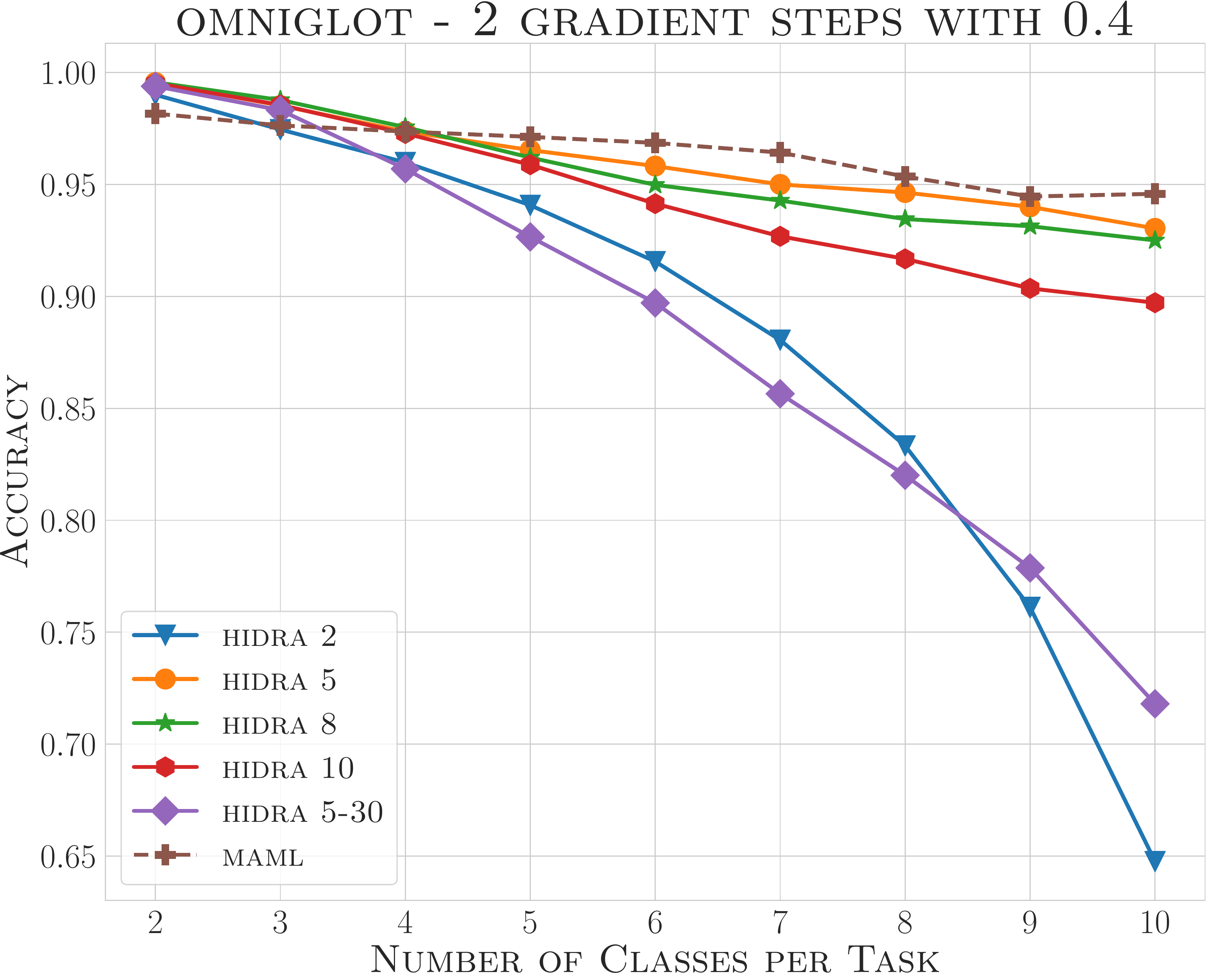}
\includegraphics[width=.32\textwidth]{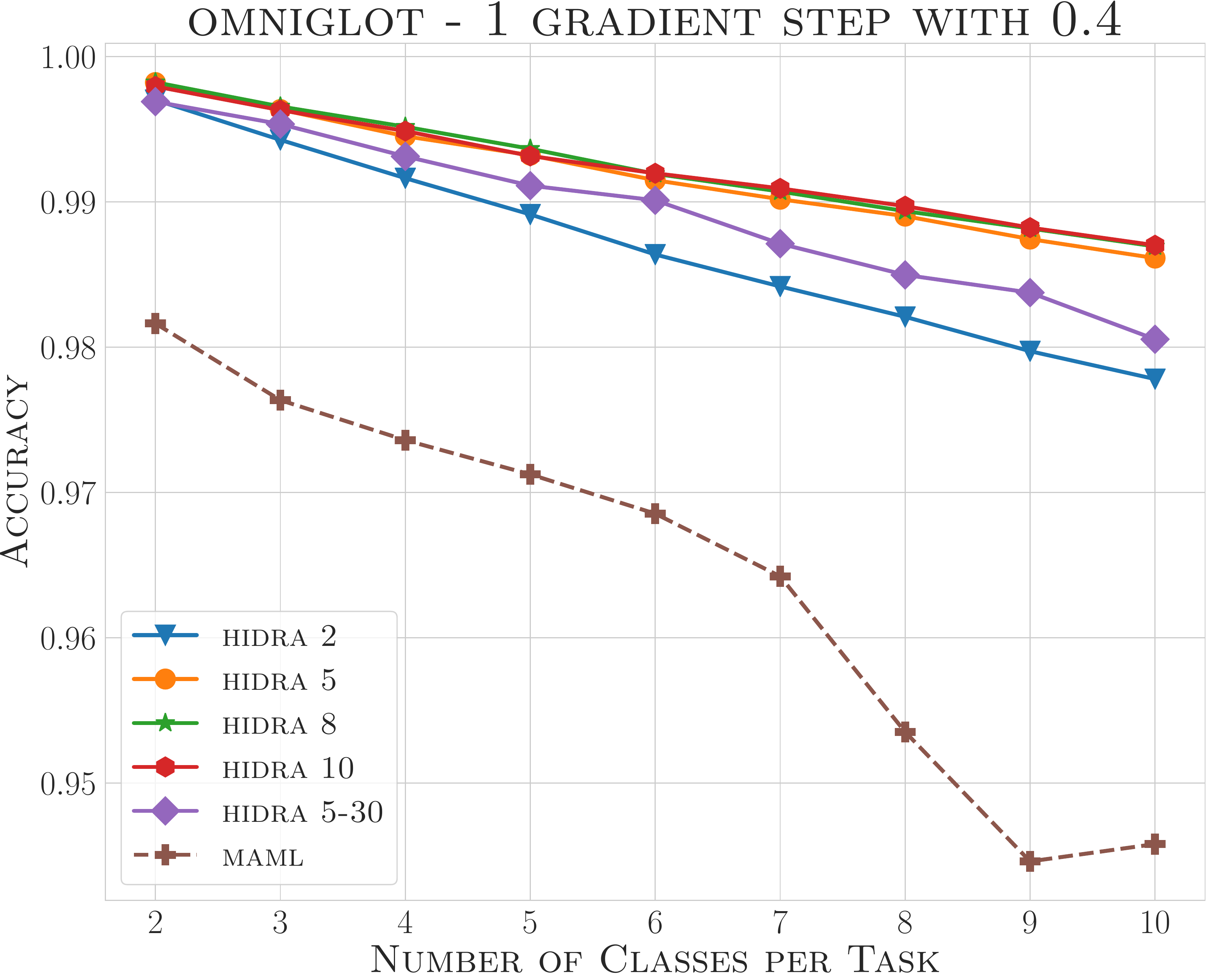}
\caption{Performance comparison between \maml{} and \hydra{} for different numbers of target variables when training on the Omniglot data set. This shows the average accuracy on tasks using each of the initializations. Each graph represent the accuracy for one amount of meta-steps. In this setup in particular we trained the \hydra{} and \maml{} experiments with learning rate equal to 0.4.} \label{omnicomp2}
\end{figure*}

%% file: plots/plotplots/OmniComp1.tex
\begin{figure*}[ht!]
\centering
\includegraphics[width=.32\textwidth]{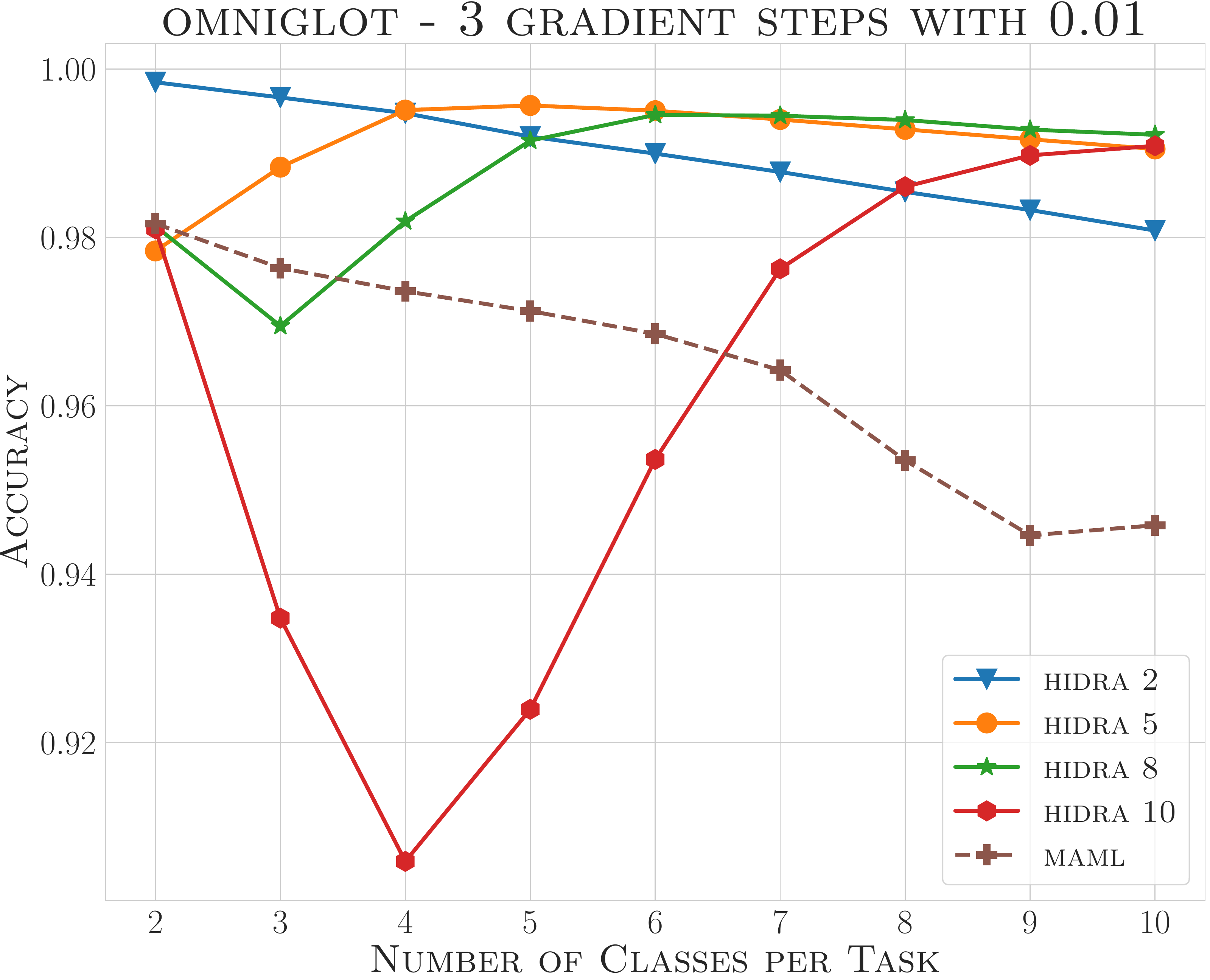}
\includegraphics[width=.32\textwidth]{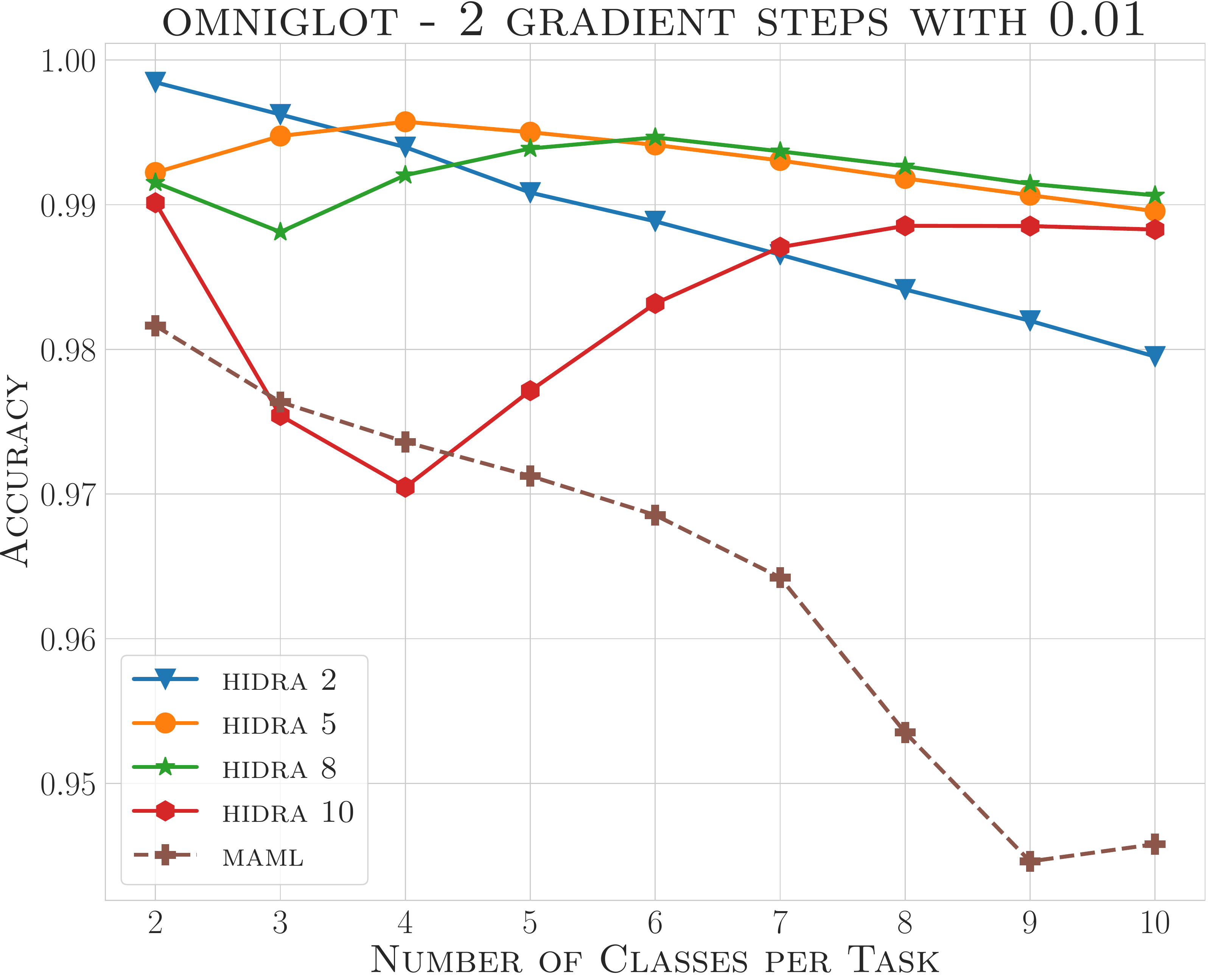}
\includegraphics[width=.32\textwidth]{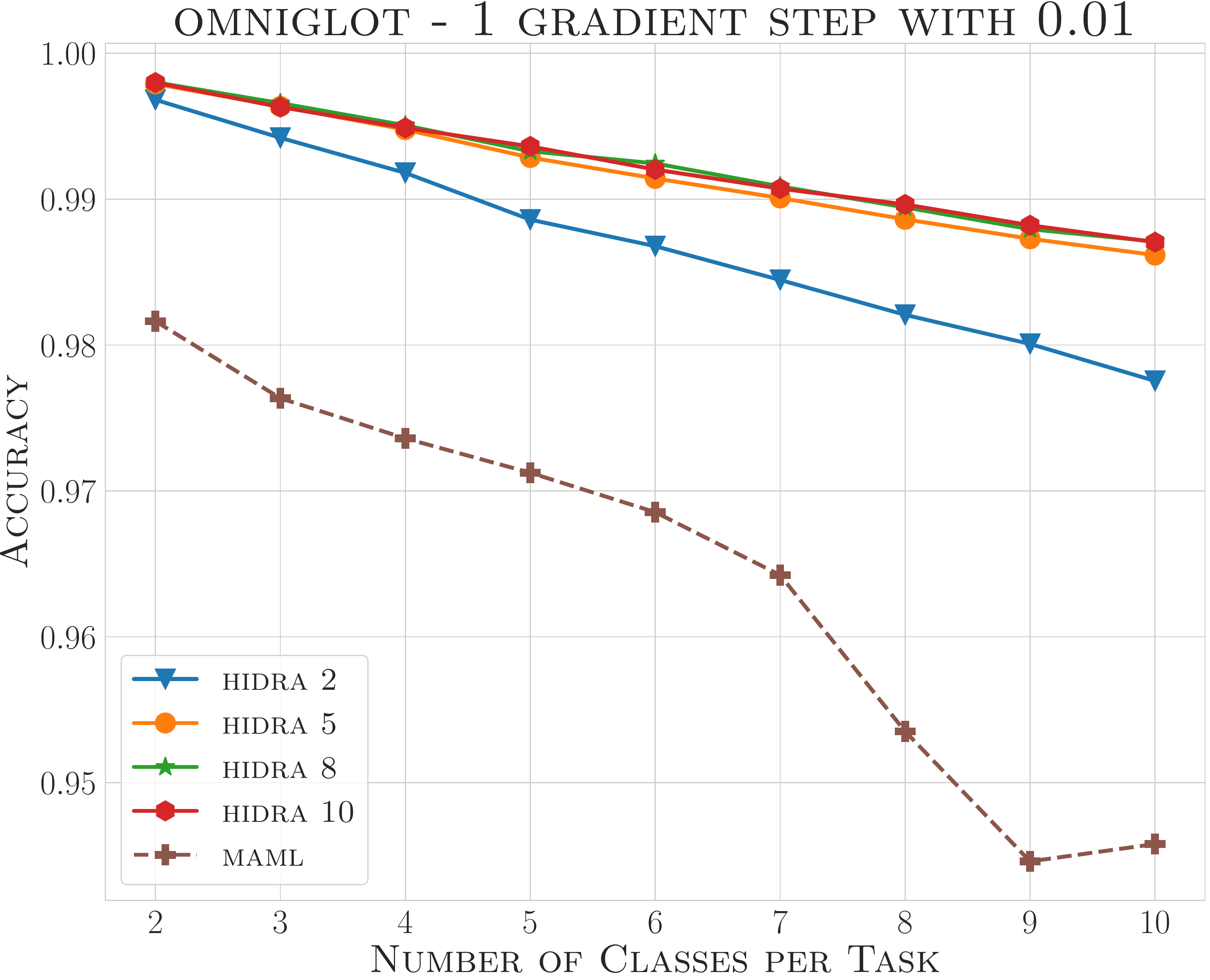}
\caption{Performance comparison between \maml{} and \hydra{} for different number of target variables when training on the Omniglot data set. This shows the average accuracy on tasks using each of the initializations. Each graph represent the accuracy for one amount of meta-steps. In this setup in particular we trained the \hydra{} experiments with learning rate equal to 0.001. \maml{} uses learning rate equal to 0.4 which remains as the best value.} \label{omnicomp1}
\end{figure*}

%% file: plots/plotplots/plotGrid.tex
\begin{figure*}[h!]
\centering
\includegraphics[width=\textwidth]{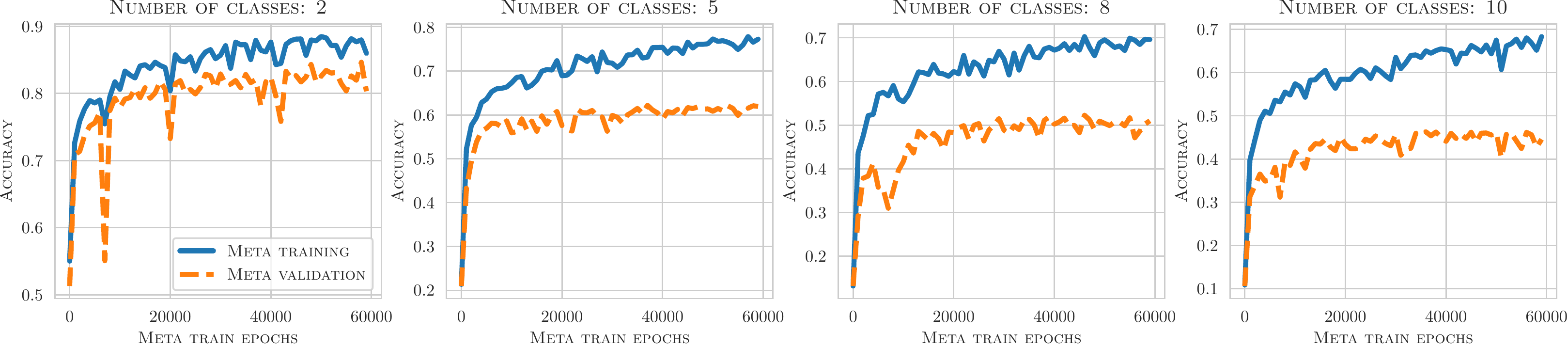}\\ ~ \\
\includegraphics[width=\textwidth]{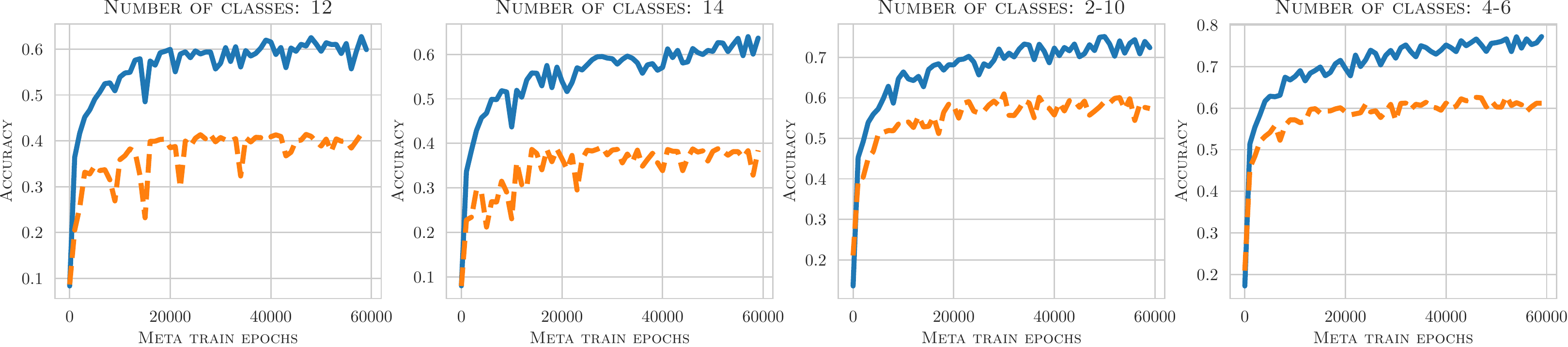}
 \
\caption{Training and validation accuracy during Meta-Learning on Omniglot displayed at every 1000th epoch.} \label{gridplot}
\end{figure*}

%% file: tables/resultsMini.tex
\begin{table*}[t!]
\centering
\adjustbox{max width=\textwidth}{
\begin{tabular}{c}
ACCURACY FOR MINIIMAGENET EXPERIMENTS \\ \midrule
\begin{tabular}{cc}

Used & Labels in Evaluation Task \\
 
\begin{tabular}{c}
Method    \\   \midrule
\textsc{hidra} 2 \\
\textsc{hidra} 5 \\
\textsc{hidra} 8 \\
\textsc{hidra} 10 \\
\textsc{hidra} 4-6 \\
\textsc{hidra} 2-10 \\
\textsc{maml} \\
\end{tabular}
&

\begin{tabular}{cccccccccccccc}
2 & 3 & 4 & 5 & 6 & 7 & 8 & 9 & 10 & 11 & 12 & 13 & 14 & 15 \\
\textbf{82.95} & 72.93 & 65.78 & 59.96 & 55.77 & 51.95 & 48.76 & 46.13 & 43.95 & 41.85 & 40.05 & 38.51 & 37.01 & 35.61 \\
77.48 & 67.58 & 60.32 & 55.21 & 50.43 & 47.25 & 44.05 & 41.54 & 39.26 & 37.54 & 35.73 & 34.51 & 33.06 & 31.8 \\
77.79 & 67.74 & 61.24 & 56.62 & 52.24 & 49.02 & 45.99 & 43.34 & 41.08 & 39.24 & 37.46 & 36.01 & 34.63 & 33.38 \\
79.94 & 68.79 & 62.44 & 57.68 & 54.27 & 51.23 & 48.36 & 45.97 & 43.32 & 41.38 & 39.43 & 37.81 & 36.37 & 34.96 \\
82.45 & \textbf{73.64} & \textbf{66.8} & \textbf{61.56} & \textbf{57.07} & \textbf{52.93} & \textbf{50.04} & \textbf{47.27} & \textbf{44.86} & \textbf{42.79} & \textbf{40.98} & \textbf{39.33} & \textbf{37.7} & \textbf{36.34} \\
82.39 & 73.2 & 66.07 & 60.73 & 56.49 & 52.38 & 49.17 & 46.7 & 44.2 & 42.36 & 40.48 & 38.88 & 37.38 & 36.03 \\
81.13 & 70.06 & 62.62 & 55.5 & 50.09 & 34.27 & 12.5 & 11.11 & 10.4 & 9.09 & 8.33 & 7.69 & 7.143 & 6.667 \\
\end{tabular}
\end{tabular}
\end{tabular}}
\caption{Average accuracy for the experiments with MiniImageNet. Each Hydra $X$ experiment (line) used $X$ amount of labels during training and was evaluated for tasks with different traget label size (columns). Hydra 2-10 and Hydra 4-6 are trained on tasks with a variable number of target variables. \maml{} is trained on a fixed output size and evaluated on the same target shape.}
\label{miexp}
\end{table*}

%% file: tables/resultsOmni.tex
\begin{table*}[t!]
\centering
\adjustbox{max width=\textwidth}{
\begin{tabular}{c}
ACCURACY FOR OMNIGLOT EXPERIMENTS\\ \midrule
\begin{tabular}{cc}

Used & Labels in Evaluation Task \\
 
\begin{tabular}{c}
Method    \\   \midrule
\textbf{Learning Rate = 0.4 1 step} \\
\textsc{hidra} 2 \\
\textsc{hidra} 5 \\
\textsc{hidra} 8 \\
\textsc{hidra} 10 \\
\textsc{hidra} 5-30 \\\midrule
\textbf{Learning Rate = 0.01 1 step} \\
\textsc{hidra} 2 \\
\textsc{hidra} 5 \\
\textsc{hidra} 8 \\
\textsc{hidra} 10 \\\midrule
\textbf{Learning Rate = 0.01 3 step} \\
\textsc{hidra} 2 \\
\textsc{hidra} 5 \\
\textsc{hidra} 8 \\
\textsc{hidra} 10 \\\midrule

\textsc{maml} \\
\end{tabular}
&

\begin{tabular}{cccccccccc}
2 & 3 & 4 & 5 & 6 & 7 & 8 & 9 & 10 \\ \midrule
\\
99.7 & 99.43 & 99.16 & 98.91 & 98.64 & 98.42 & 98.21 & 97.97 & 97.78 &  \\
99.82 & 99.64 & 99.45 & 99.32 & 99.15 & 99.02 & 98.9 & 98.74 & 98.61 &  \\
99.82 & 99.66 & 99.52 & 99.37 & 99.19 & 99.07 & 98.94 & 98.82 & 98.69 &  \\
99.8 & 99.63 & 99.49 & 99.32 & 99.2 & 99.09 & 98.97 & 98.82 & 98.7 &  \\
99.69 & 99.54 & 99.31 & 99.11 & 99.01 & 98.71 & 98.5 & 98.38 & 98.05 &  \\\midrule

\\
99.68 & 99.42 & 99.18 & 98.86 & 98.68 & 98.45 & 98.21 & 98.01 & 97.75 &  \\
99.79 & 99.64 & 99.48 & 99.29 & 99.14 & 99.01 & 98.86 & 98.73 & 98.62 &  \\
99.8 & 99.66 & 99.51 & 99.33 & 99.24 & 99.09 & 98.94 & 98.79 & 98.71 &  \\
99.8 & 99.63 & 99.49 & 99.36 & 99.2 & 99.07 & 98.96 & 98.82 & 98.71 &  \\ \midrule

\\%
\textbf{99.85} & \textbf{99.69} & \textbf{99.54} & 99.31 & 99.13 & 98.91 & 98.69 & 98.48 & 98.24 &  \\
97.05 & 97.64 & 99.17 & \textbf{99.57} & \textbf{99.53} & \textbf{99.46} & 99.37 & 99.28 & 99.19 &  \\
97.73 & 96.13 & 96.89 & 98.27 & 99.22 & 99.43 & \textbf{99.42} & \textbf{99.33} & \textbf{99.3} &  \\
98.09 & 91.89 & 87.72 & 87.94 & 89.96 & 92.81 & 95.53 & 97.65 & 98.66 &  \\ \midrule

98.16 & 97.64 & 97.36 & 97.12 & 96.85 & 96.42 & 95.35 & 94.46 & 94.58 \\

\end{tabular}
\end{tabular}
\end{tabular}}
\caption{Average accuracy for the experiments with Omniglot. Each Hydra $X$ experiment (line) is trained with tasks containing $X$ target variables and evaluated for tasks with different label size (columns). Hydra 5-30 is trained on tasks with a variable number of target variables. \maml{} is trained and evaluated on a fixed output size.}
\label{omexp}
\end{table*}

%% file: chapters/6-conclusion.tex
\section{Conclusion} \label{sec:Conclusion}

In this paper, we present a novel approach for learning a task-specific initialization through meta-learning. We show that while \maml{} is capable of learning such an initialization, it is restricted to a fixed number of classes while including redundancy in the learned output layer, which is demonstrated to hinder learning across tasks with a high number of classes when using a low-capacity model. \hydra{} solves both of these problems by training a single master neuron, which is used to initialize output neurons dynamically.
Experiments on common few-shot classification benchmarks demonstrate that a single \hydra{} model can generalize to all number of classes independent of the number of target variables used during meta-learning. At the same time, this is shown to lead to a more robust architecture that is able to train on tasks with a high number of classes, where \maml{} is not applicable.
This technique however is currently limited to the output space and does not work with latent or input spaces.
Finally, using a single model initialized with \hydra{} is shown to improve on the results achieved with a set of models initialized with fixed output layers.